\pdfoutput=1
\documentclass[11pt]{article}

\usepackage{times}
\usepackage{latexsym}

\usepackage[T1]{fontenc}
\usepackage[utf8]{inputenc}

\usepackage[colorlinks=true,citecolor=blue,urlcolor=cyan]{hyperref} 
\usepackage[authoryear,round,longnamesfirst]{natbib}
\usepackage{graphicx}

\usepackage{makecell}
\usepackage{booktabs}

\title{PeLLE: Encoder-based language models for Brazilian Portuguese based on open data}

\author{
	\hspace*{-2em}
	\begin{tabular}{c@{~~~~~~~~~~~~~}c}
		\multicolumn{2}{c}{
			Guilherme Mello, Marcelo Finger, Felipe Serras, Miguel de Mello Carpi, Marcos Jose} \\
		\multicolumn{2}{c}{	University of São Paulo }
  \\[3ex] \hspace*{2em}
	  Paulo Cavalin & Pedro Henrique Domingues \\
	  \hspace*{2em}
	  IBM Research &
	  IBM Research, PUC-Rio 
	\end{tabular}
}

\begin{document}
\maketitle

\begin{abstract}
    In this paper we present PeLLE, a family of large language models based on the RoBERTa architecture, for Brazilian Portuguese, trained on curated, open data from the Carolina corpus. Aiming at reproducible results, we describe details of the pretraining of the models.  We also evaluate PeLLE models against a set of existing multilingual and PT-BR refined pretrained Transformer-based LLM encoders, contrasting performance of large versus smaller-but-curated pretrained models in several downstream tasks. We conclude that several tasks perform better with larger models, but some tasks benefit from smaller-but-curated data in its pretraining.
\end{abstract}

%======================================================================
\section{Introduction}
\label{sec:intro}

We introduce PeLLE, a family of Brazilian \underline{P}ortugues\underline{e} \underline{L}arge \underline{L}anguage \underline{E}ncoder, and examine different ways of Large Language Model pretraining. We evaluate its performance in downstream NLP tasks for Brazilian Portuguese (BP).  The goal of this work is to produce and evaluate BP-specific versions of a Transformer-encoded language model~\citep{vaswani2017attention} based on the RoBERTa architecture~\citep{liu2019roberta}, in a reproducible way.  We investigate whether language-specific training is preferable to extending multilingual models, such as mBert~\citep{devlin2018bert} and XLM-R~\citep{conneau-etal-2020-unsupervised} with training over the same open corpus.

The use of an open corpus ensures reproducible comparisons, and for that we choose Carolina Corpus~\citep{ramos2023carolina}, an open general corpus of Brazilian Portuguese.  Furthermore, Corpus Carolina is a highly curated corpus, as far as very large corpora can be curated, which also allows us to contrast the employment of quality data versus non-curated big data.

Model evaluation is performed over nontrivial NLP tasks for Portuguese which still challenge existing technology, such as natural language inference, hate speech identification and classification. We also add to the comparison several baseline models derived from existing BP LLM-encoders, such as Bertimbau~\citep{souza2020bertimbau} and Albertina-PTBR~\citep{rodrigues2023advancing}, and generic multilingual ones, such as mBERT~\citep{devlin2018bert} and XLM-R~\citep{conneau-etal-2020-unsupervised}.

All our pretrained models will be made publicly available on Huggingface.

%======================================================================
\section{Related work}
\label{sec:related_work}

Large Language Models (LLMs) have emerged in the past few years as an effective way to improve Neural Network (NN)-based models for Natural Language Processing (NLP)-based applications \citep{devlin2018bert,brown2020}. The main idea consists of \emph{pretraining} a NN on a large set of data relying on self-supervised training objectives. That is, based on Auto-Regressive or Masked Language Modeling (MLM) training objectives. A commonly-used approach in current days is to first \emph{pretrain} a Transformer NN on a large and generic set of unlabeled data, and then \emph{finetune} its parameters to a more specific and smaller dataset with labeled data. We usually refer to those datasets as \emph{pretraining data} and \emph{downstream data} \citep{raffel2020}. It is worth adding that LLMs can be based on different types of Transformer-based architectures, such as those relying on a full Encoder-Decoder architecture \citep{raffel2020}, or those based on either Encoder \citep{devlin2018bert} or Decoder \citep{brown2020} sides. In this work we focus specifically on Encoder-based LLMs. 

It is well known that English is the language with most abundant data in our current day's digital world, and one can easily scrape a lot of data from the web to pretrain very large models \citep{Joshi2020stateFate}. For this reason, it is natural that most recent efforts in LLMs were targeted at English at first, and other less-resourced languages have a much smaller coverage of the LLM advances. Taking Portuguese as an example, since is the target language of this work, we can observe that it is included in some multilingual models \citep{devlin2018bert,xue-etal-2021-mt5}, and there are some specific efforts in building Portuguese-only models \citep{souza2020bertimbau,rodrigues2023advancing}. 

Among the Portuguese specific models, the main difference is usually the pretraining data. Besides the fact that the pretraining data can impact the performance of the model on downstream data, it is also very important to understand the licenses of the documents that were used to compose that dataset. One corpus that is commonly used is BrWaC~\citep{wagner2018brwac}, which contains web pages crawled with the help of a web search engine. Although it is common belief that publicly-available content can be freely used without asking for permission, that is not true. Each web page has specific copyright licenses and it is key to understand such licenses before using the data to pretrain an LLM. That said, we can find models such as Bertimbau and Albertina-PTBR that are pretrained on BrWaC, and that makes these models unsuitable for commercial applications. That is the reason for the release of an alternative version of Albertina-PTBR, namely Albertina-PTBR-NoBrWaC, pretrained on data with more permissive licenses such as the OSCAR dataset \citep{abadji-etal-2022-towards}.

Our work proposes a model that is pretrained only on complete open data in Brazilian Portuguese, Carolina Corpus, which is described in section \ref{sec:carolina}. This allows us to present a model that can be reproduced by being totally transparent in terms of data licensing. 
Although we present models pretrained on the full dataset, future models could use only specific documents given licenses of choice. Additionally, the taxonomies provided in the corpus, i.e. the classification of each document according to a set of specific domains, allows for training models for specific domains in the future.

%======================================================================
\section{Carolina Corpus}
\label{sec:carolina}

The Carolina Corpus\footnote{\url{https://sites.usp.br/corpuscarolina/corpus/}} is a general corpus of Brazilian Portuguese in continuous development. The iteration employed for the training of our models corresponds to version 1.2 Ada, which was the latest version available during the pretraining phase. The corpus was constructed with the dual purpose of allowing investigations in linguistics and in artificial intelligence, particularly in the pretraining of large language models, as exemplified in the present study \citep{ramos2023carolina}.

Version 1.2 of the Carolina Corpus has been constructed through the extraction of texts from the Web. Its primary distinguishing feature, in comparison to similar corpora, is the systematic focus on both provenance and typological diversity of included texts.

The Corpus has been constructed employing the WaC-WiPT methodology (Web-as-Corpus With Provenance and Typology) \citep{sturzeneker2022carolina}, a variation of the prevailing Web-as-Corpus (WaC) approach \citep{baroni2006wacky,baroni2009wacky}. This method encompasses a sequence of stages for systematically exploring and curating targeted web domains prior to their crawling and data extraction. Such procedure ensures the incorporation of only open-licensed texts, as well as the recording of diverse metadata, encapsulated within the header of each text. These encompass the original license of the text, responsible organization and various typological information.

The version 1.2 of Carolina Corpus has 823 million words, segmented across more than 2 million texts. In terms of typological diversity, these tokens are mainly distributed in five types: instructional, juridical, virtual (including social media content), entertainment, and journalistic texts \citep{slides_carolina12}.

As highlighted in Table~\ref{tab:datasets_comparision}, Corpus Carolina stands out as being the more open option when comparing to the datasets used to train other Portuguese encoders. The methodology adopted in its development guarantees the addition of permissive license texts and maintains information about the original license of each text. This allows the corpus and the models derived from it to be more freely distributed and used.

\begin{table*}[ht!]
    \centering
    \caption{Comparision of Datasets used to train Brazilian Portuguese models.}
    \label{tab:datasets_comparision}
    \resizebox{\textwidth}{!}{
        \begin{tabular}{cccm{7cm}}
            \toprule
                 \textbf{Model Family} & \textbf{Training Dataset} & \textbf{Size (tokens)} & \textbf{Openness} \\
            \midrule
            \makecell{BERTimbau\\Albertina-PTBR} & BrWaC & 2.68B & 
                 Crawled from the web, available for research purposes only, download upon request \\
             \midrule
                 Albertina-PTPT & Oscar (mainly) & 2.20B & 
                 Crawled from the web, shuffled at line level to avoid copyright issues, available in public domain, download upon request \\
            \midrule
                 PeLLE & Carolina v1.2 & 0.82B & 
                 Crawled by type from  curated web-domains, free download, licenses specified for each text. \\
             \bottomrule
        \end{tabular}
    }
\end{table*}

%======================================================================
\section{PeLLE models}
\label{sec:pelle}

In this section we present the details on how we created the PeLLE family of LLMs introduced in this work. We present three new models, namely \emph{pPeLLE}, \emph{xPeLLE}, and \emph{mPeLLE}, based on RoBERTa~\citep{liu2019roberta}, XML-R~\citep{conneau-etal-2020-unsupervised}, and mBERT~\citep{devlin2018bert}, respectively. Notice that our three models were trained on the same 1.2 version of the Carolina corpus.

The pPeLLE model is based on the RoBERTa~\textit{base} model and was trained from scratch, using Masked Language Modeling (MLM) objective, with a new vocabulary using the Carolina Corpus (v1.2) only. Even though DeBERTa is consistently better than RoBERTa \citep{He2020deberta}, as literature presents, model performance can be mainly improved by model and data size rather than specific improvements in model architecture or attention mechanism \citep{Kaplan2020scaling, Tay2022scale}. Thus, we used RoBERTa for it is a simple and robust method for transformer's  pretraining and left DeBERTa as a future work. To develop the new vocabulary we followed the RoBERTa tokenization and used a Byte-Level Byte-Pair Encoding with 50k sub-word units. As Portuguese makes use of accents, we used NFKD Unicode normalization as the only text preprocessing step.

To chose the pretraining hyperparamenters, we first conducted a brief search, varying the learning rate: \{1e-4, 1e-3\}; and batch size \{2k, 4k, 8k\}. To keep the training of each model similar in number of total tokens, the maximum training steps were adjusted according to the batch size to reach 40 epochs of training, to keep an approximate training time (wall clock). Similar to RoBERTa, and following the literature, using bigger batches results in better performance (lower loss) \citep{Izsak2021bert}. Using the best hyperparameters (lr: 1e-4; bsz: 8k) the PeLLE model was trained for 100k steps.

To evaluate the impact of multilingual models, we used mBERT and XLM-R weights to initialize mPeLLE and xPeLLE, respectively. We kept they original multilingual vocabulary and used the same pretraining hyperparameters as PeLLE. xPeLLE was trained for 45k steps while mPeLLE was trained for 20k steps, also using the Carolina Corpus v1.2.

All three models are available on Huggingface (links not provided in this version for anonymity reasons).

%======================================================================
\section{Evaluation}
\label{sec:evaluation}

In this section we present the evaluation of models on downstream tasks. For that, we rely on a varied set of benchmark datasets for Brazilian Portuguese. More specifically, we apply our models and baselines on the ASSIN, ASSIN 2, HateBR, and Acórdãos TCU datasets. We believe this evaluation protocol helps us understand the capability of our models, not only on different types of data but also on different types of tasks.

For the aforementioned datasets, we conduct a full finetuning of the models, where the classification layer used for MLM prediction is replaced by a new layer for the downstream task. Then, we train this new layer and adjust the remaining parameters for the new task. For that we rely on the HuggingFace library, \textit{transformers}\footnote{https://github.com/huggingface/transformers}, which provide us the base coding structure to train and evaluate the models.

Notice that, in terms of implementation, the main modification for each task lies in setting up a different top-most layer for the model, according to the number of classes involved in the problem or the type of task itself. If it is a regression task, we need a single-value layer for predicting the final output, and if it is a classification problem, the neural network needs a layer with as many outputs as the number of classes. In addition, we also need to use the appropriate training loss function, which is the Cross Entropy loss for classification and Mean Squared Error (MSE) for regression.

Next, we describe the baseline models, experiments and results on each downstream dataset.

%----------------------------------------------------------------------
\subsection{Baseline models}
\label{sec:baseline_models}

To compare the performance of the proposed models, we consider a set LLMs pretrained either on Brazilian Portuguese corpora or multilingual datasets. They are presented as follows.

% BERTimbau
\emph{\textbf{BERTimbau}}~\citep{souza2020bertimbau} is from the family of BERT models developed for Brazilian Portuguese. Pretrained with BrWaC dataset, the largest Brazilian Portuguese corpus, comprising 2.68B tokens from documents of several domains, BERTimbau achieved the state-of-the-art for Named Entity Recognition (NER), Sentence Textual Similarity (STS) and Recognizing Textual Entailment (RTE) tasks in Brazilian Portuguese at the time it was released. For this work, we consider both BERTimbau-\emph{Base} and BERTimbau-\emph{Large} models.

% AlBERTina
\emph{\textbf{Albertina}}\footnote{We emphasize that Albertina-PTBR-Base was removed from their repository at the time of development of this paper and may not be available for further analysis.} \citep{rodrigues2023advancing} is another family of LLMs designed for Brazilian and European Portuguese. These models are based on DeBERTa \citep{He2020deberta}, which was able to surpass humans on the SuperGLUE benchmark. In this paper, we consider only the Brazilian Portuguese models: \emph{Albertina-PTBR-Base}, \emph{Albertina-PTBR-Large}, and \emph{Albertina-PTBR-NoBrWaC}. While the pretraining data for Albertina-PTBR-base and Albertina-PTBR-Large were based on BrWaC, Albertina-PTBR-NoBrWaC relied on a set of documents from the OSCAR dataset, making it a model with less-restricted licenses in pretraining data.

% mBERT
\emph{\textbf{mBERT}}\footnote{https://huggingface.co/bert-base-multilingual-cased} is the multilingual version of BERT \textit{base} \citep{devlin2018bert}. It was pretrained with data from 104 different languages, considering the languages with largest Wikipedia representation.

% XLM-RoBERTa
\emph{\textbf{XLM-RoBERTa}} (XLM-R) is built upon the RoBERTa model architecture and is pretrained on a large corpus containing 100 different languages, which enables it to perform well on a variety of cross-lingual tasks without the need for language-specific training data \citep{conneau-etal-2020-unsupervised}. 

% Debertinha
% No fim eu acredito que não vale a pena citar, já que apenas uma métrica coincide com o
% que a gente mediu (ASSIN2-STS) 84.74

%----------------------------------------------------------------------
\subsection{ASSIN and ASSIN 2}
\label{sec:assin_dataset}

 ASSIN \citep{fonseca2016assin} and ASSIN 2 \citep{real2020assin} are datasets for the tasks of Recognizing Textual Entailment (RTE) -- also known as Natural Language Inference -- and Semantic Textual Similarity (STS). They contains up to 10,000 pair of sentences annotated with gold standard. STS task is a regression problem with a target score ranging from 1 to 5. RTE is a classification problem but ASSIN and ASSIN 2 have different label sets. ASSIN RTE task is a multi-class classification with sentences labeled as \textit{entailment}, \textit{paraphrase} or \textit{none}. The ASSIN 2 RTE task was simplified to a binary classification problem with target labels only stating if the entailment holds or not (\textit{entailment} or \textit{none}).

Each dataset is divided into training, validation and test splits but their distribution are different (see Table~\ref{tab:assin_dataset}). While ASSIN contains sentences in Brazilian and European Portuguese, equally distributed across each split, ASSIN 2 contains only the Brazilian variation.

\begin{table}[!h]
    \centering
    \caption{ASSIN and ASSIN 2 dataset splits.}
    \label{tab:assin_dataset}
    \begin{tabular}{lrrrr}
            & Train & Val. & Test  & \textbf{Total}  \\
    \hline
    ASSIN   & 5,000 & 1,000      & 4,000 & \textbf{10,000} \\
    ASSIN 2 & 6,500 &   500      & 2,448 & \textbf{ 9,448}
    \end{tabular}
\end{table}

To evaluate each model in this downstream tasks, we performed a hyper-parameter search using a Population Based Training (PBT) \citep{Jaderberg2017pbt}. We searched over the batch sizes of 16, 32 and 64 and learning rates 5e-6, 5e-5, 5e-4 and 5e-3, resulting in 12 initial models (agents) trained for 10 epochs with perturbation between each epoch.

\begin{table*}[ht!]
    \centering
    \caption{ASSIN and ASSIN 2 results. RTE task are measured by F1 Score and STS by Pearson Correlation.}
    \label{tab:assin_results}
    \begin{tabular}{lccccc}
                & \multicolumn{2}{c}{ASSIN}
                & \multicolumn{2}{c}{ASSIN 2} \\
                \cmidrule(r){2-3} \cmidrule(l){4-5}
    Model       & RTE    & STS    & RTE    & STS & \textit{avg} \\ \toprule
    Albertina PTBR Large    & \textbf{0.8959} & \textbf{0.8746} & \textbf{0.9152} & \textbf{0.8680} & \textbf{0.8884} \\
    Albertina PTBR NoBrWaC  & 0.8943 & 0.8693 & 0.9040 & 0.8543 & 0.8804 \\
    Bertimbau Large         & 0.8613 & 0.8532 & 0.8891 & 0.8495 & 0.8632 \\ \hline
    Albertina PTBR Base     & 0.8292 & 0.8237 & 0.8762 &   --   & 0.8430 \\
    Bertimbau Base          & \textbf{0.8486} & \textbf{0.8423} & \textbf{0.8883} & \textbf{0.8443} & \textbf{0.8558} \\
    mBERT                   & 0.8052 & 0.7982 & 0.8373 & 0.8163 & 0.8142 \\
    XLM-R Base              & 0.8148 & 0.8192 & 0.8703 & 0.8026 & 0.8267 \\
    pPELLE                  & 0.8113 & 0.8042 & 0.8489 & 0.8059 & 0.8175 \\
    mPELLE                  & 0.7818 & 0.7981 & 0.8437 & 0.7942 & 0.8044 \\
    xPELLE                  & 0.8184 & 0.8294 & 0.8551 & 0.8232 & 0.8315 \\
    \bottomrule
    \end{tabular}
\end{table*}

The results in Table~\ref{tab:assin_results} shows that larger models results in better performance, which is in accordance with the literature \citep{Kaplan2020scaling}, but we focus our analysis on smaller (base) models as they represent a larger variability in its method of construction. As BERTimbau performs better among \textit{base} models, it is possible to assign its improved performance to its pretraining design. BERTimbau \textit{base} used mBERT weights as starting point. Also, they retrained the model tokenizer from scratch, creating a new vocabulary, which is known to improve model performance \citep{Artetxe2020cross-lingual}. pPeLLE performance is another evidence for that as it performs close to multilingual models and was trained with a new vocabulary. When comparing PeLLE family models with multilingual models, we can observe that multilingual models is a good choice for weight initialization and XLM-R is also a better choice for Portuguese as xPeLLE stands out among PeLLE family.

%----------------------------------------------------------------------
\subsection{HateBR}
\label{sec:hatebr_dataset}

HateBR is an annotated dataset of offensive language on social media data \citep{vargas-etal-2022-hatebr}. It is comprised of 7,000 Instagram\footnote{https://www.instagram.com} comments annotated according to three distinct layers: 1) binary classification in offensive versus non-offensive content; 2) multi-class classification in nine hate speech groups, i.e. xenophobia, racism, homophobia, sexism, religious intolerance, partyism, apology for the dictatorship, antisemitism, and fatphobia; and 3) a regression problem to compute offensiveness levels into highly, moderately, and slightly offensive messages. We believe that this dataset not only allows us to evaluate the LLMs into a real-world application with real-world data, but also provides an interesting range of applications for evaluating the models, i.e. binary classification, multiclass classification, and regression.

We split the 7,000 samples of the dataset in 80\% of the examples for training, 10\% for validation, and 10\% for test. The selection of samples was random and stratified considering the distribution of labels for the binary classification task\footnote{The data splits will be available online}, and we relied on the same split for the regression. Notice that the binary problem is almost balanced, with 3,500 samples of non-offensive texts and 3,477 with offensive ones, so our data splitting aimed at keeping that balance. For the multiclass problem, we also considered the stratification of the dataset considering the distribution of the classes. Since the multiclass problem is also a multilabel problem, where a sample can be associated to more than one class at once, we simplified the implementation by computing the set with all possible combinations of labels and considered each combination as a separate class. Given that some classes presented too few example for a proper division into training, validation, and test sets, classes with less that 10 samples were discarded. 
The list of classes and respective number of samples is presented in Table~\ref{tab:hatebr_class_distribution}.

\begin{table}
    \centering
    \caption{HateBR class distribution. }
    \label{tab:hatebr_class_distribution}
    \begin{tabular}{lc}
         Class                          & \#samples \\
    \toprule
        non-offensive                   &  3500 \\
        offensive \& non-hate speech    &  2798 \\
        apology for the dictatorship    &    28 \\
        fatphobia                       &    20 \\
        homophobia                      &    15 \\
        partyism                        &   479 \\
        partyism AND sexism             &    11 \\
        religious intolerance           &    46 \\
        sexism                          &    80 \\
    \bottomrule
    \end{tabular}   
\end{table}

It is also worth mentioning that, for these evaluations, we adopted a simplified set of experiments. All models were trained with the same set of meta-parameters, i.e. learning rate set to 5e-5, batch size set to 16, and number of epochs set to 5, and we consider only the baselines with comparable sizes. The main idea is to understand how the models compare in a fast-deployment setting, for which computation resources are not abundant and one needs to train smaller models with limited optimizations. Thus, the models we considered as baselines are mBERT, Bertimbau-Base and Albertina-PTBR-Base, where the latter two models are simply referred to as Bertimbau and Albertina-PTBR in this section.

In Table~\ref{tab:hatebr_tcu_results} (see \emph{2 Classes} column), we present the F1 scores for the binary classification task. We observe that all models perform quite close in this task. pPeLLE and xPeLLE achieves the same values of Bertimbau, with 0.91, and these three models are slightly better than mPeLLE and Albertina-PTBR, both with an F1 score of 0.90. The multilingual mBERT model is slightly worse than the top performers, with 0.87.

\begin{table*}[ht!]
    \centering
    \caption{Results for the HateBR and \textit{Acórdãos} TCU tasks. The results of the \textit{Acórdãos} TCU datasets are presented in the "TCU dataset" column and are expressed in terms of F1 Score. For HateBR, the binary classification (2 Classes column) and multiclass classification (9 Classes column) are also expressed in terms of F1 Score. The regression taks in HateBR dataset are expressed in terms of $R^2$.}
    \label{tab:hatebr_tcu_results}
    \begin{tabular}{lccccc}
        & \multicolumn{3}{c}{HateBR} &   \\
    \cmidrule{2-4}
    Model           & 2 Classes      & 9 Classes    & Regression    & TCU Dataset   & \textit{avg} \\
    \toprule
    pPeLLE          & \textbf{0.91} & 0.47          & 0.47          & \textbf{0.82} & 0.6675 \\
    xPeLLE          & \textbf{0.91} & 0.43          & 0.50          & 0.77          & 0.6525 \\
    mPeLLE          & 0.90          & 0.49          & 0.49          & 0.81          & 0.6725 \\
    \midrule
    mBERT           & 0.87          & 0.39          & 0.43          & 0.72          & 0.6025 \\
    Bertimbau       & \textbf{0.91} & \textbf{0.50} & \textbf{0.54} & 0.80          & \textbf{0.6875} \\
    Albertina-PTBR  & 0.90          & 0.47          & 0.45          & \textbf{0.82} & 0.6600 \\
    \bottomrule
    \end{tabular}
\end{table*}

The results with the multi-class classification are presented in Table~\ref{tab:hatebr_tcu_results} (see \emph{9 Classes} column). Here we observe a slightly different scenario from the binary classification. In this problem Bertimbau performed best with an F1 score of 0.50, followed by mPeLLE, with 0.49, and by pPeLLE and Albertina-PTBR, both with F1s of 0.47. Once again, mBERT performed the worst, with an F1 of only 0.39. And xPeLLE presented modest results, with an F1 score of 0.43.

The results for the regression task are presented in Table~\ref{tab:hatebr_tcu_results} (see \emph{Regression} column). First, it is worth mentioning that the models present quite modest $R^2$ scores, with the best result achieved by Bertimbau of only 0.54. The PeLLE models were closest competitors, with 0.50, 0.49, and 0.47 for xPeLLE, mPeLLE, and pPeLLE, respectively. The Albertina-PTBR was slightly worse, with 0.45, followed by mBERT with 0.43, which was the worst model once again.

From these results, the first main conclusion is that the use of open data seems not to affect so much the performance of the models when finetuned to a downstream task. In the binary classification task, pPeLLE achieves the best result, and at least one of our three models was a strong contender in the other two tasks. This is encouraging since models pretrained on open data can be used with no restriction in terms of data. Additionally, the not-so-stellar results of Albertina-PTBR indicate that a smaller pretraining dataset can result in better models if the data is well-curated, which is the case of the Caroline corpus. Finally, it seems clear that Portuguese specific models compose a better choice for these tasks, given that mBERT usually performs worse than the Portuguese-specific models.

%----------------------------------------------------------------------
\subsection{Acórdãos TCU dataset}
\label{sec:tcu_dataset}

We complement the evaluation of the models on the case-law documents from the Acórdãos TCU dataset\footnote{https://www.kaggle.com/datasets/ferraz/acordaos-tcu}. This dataset is particularly interensting to evaluate whether the legal documents included in Carolina help the model to better handle legal-domain documents. In addition, this dataset is also useful to evaluate the models on longer documents, since the average document length is 225.75 words, with document lengths ranging from 8 to 4929 words. For comparison purposes, the mean length of the texts in the HateBR dataset is of 11.56 words.

For this evaluation we focused on the classification of the type of the document, consisting of 29 classes. This dataset contains 35,414 samples, with an average number of examples per class of 1,221.55, with the number of examples per class ranges from 20 to 12,659. Since there are classes that we consider with too few examples, such as single-digit number of examples, we removed such low-resourced classes and considered only the remaining 24 classes to train our models. 

For the experimental setup, we followed a similar training approach to that with HateBR, described in Section~\ref{sec:hatebr_dataset}. We considered the same meta-parameters and the same baseline models, i.e. mBERT, Bertimbau-Base, and Albertina-PTBR-Base.

The results in terms of macro F1 scores are presented in Table~\ref{tab:hatebr_tcu_results} (see \emph{TCU Dataset} column). We observe that pPeLLE achieves the highest F1 score with 0.82, tied with Albertinha-PTBR. The mPeLLE and Bertimbau models presented values that were close to that, i.e. 0.81 and 0.80, respectively. In addition to that, xPeLLE was slightly worse with 0.77. Again, mBERT lagged behind the other models, with a modest score of only 0.72. 

These results somehow confirm out previous outcomes. That is, the models from the PeLLE family usually present competitive results compared with models trained on larger and non-curated datasets, such as Bertimbau and Albertina-PTBR. Given that there is one PeLLE model, i.e. pPeLLE, that achieved the best performance on this dataset, we may claim that the use of law-related documents in the pretraining data can result in a model that excels on legal-domain downstream tasks. Off course, further experiments with other legal-domain datasets are necessary to make such claim stronger.

%======================================================================
\section{Conclusions and Future Steps}
\label{sec:conclusions}

The results of the evaluation above show that several tasks performs better with larger models, but some benefits from smaller-but-curated data in its pretraining. We do not have a clear distinction between those two sets of tasks, in general.  

However, for tasks involving regression, as in Section~\ref{sec:hatebr_dataset}, because all evaluated models are pretrained for classification only, larger models clearly have an advantage. So size demonstrably dominates when all else is equalized.

It is also clear that pure Portuguese-only pretrained model, pPeLLe, for the size of curated corpus used in this work, around 0.85B tokens, is sometimes a competitive performer in Portuguese-only tasks against mPeLLE and xPeLLE.

Models of the PeLLE family that extend the weights of multi-language models such as mBERT and XLM-R are often quite competitive, usually just a few percentage points below the best performing Large models. Neither xPeLLe nor mPeLLE is a clear best performer, but the former has come up front more frequently.

The points above reinforce the advantage of continuing the enlargement of the Carolina is a Corpus, which describes itself as in constant development.  Its future versions may serve the basis for larger models which may benefit both from the size and quality of curated data.  On top of allowing for reproducible experiments due to the employment of open datasets,  this new training will hopefully allow us to better understand the impact of the amount of training data on model performance.

%What is the loss compared to non-open models?

%Vale a pena treinar do zero ou não? future work

%As Carolina is a Corpus in constant development, future versions, which include the addition of significant volumes of textual data, may allow the training of new versions of our models. This new training will allow us to better understand the impact of the amount of training data on model performance.

%\bibliographystyle{dinat}  
\bibliography{bibliography}  

\begin{thebibliography}{}

\bibitem[Abadji et~al., 2022]{abadji-etal-2022-towards}
Abadji, J., Ortiz~Suarez, P., Romary, L., and Sagot, B. (2022).
\newblock Towards a cleaner document-oriented multilingual crawled corpus.
\newblock In {\em Proceedings of the Thirteenth Language Resources and
  Evaluation Conference}, pages 4344--4355, Marseille, France. European
  Language Resources Association.

\bibitem[Artetxe et~al., 2020]{Artetxe2020cross-lingual}
Artetxe, M., Ruder, S., and Yogatama, D. (2020).
\newblock On the cross-lingual transferability of monolingual representations.
\newblock In {\em Proceedings of the 58th Annual Meeting of the Association for
  Computational Linguistics}, pages 4623--4637. Association for Computational
  Linguistics.

\bibitem[Baroni et~al., 2006]{baroni2006wacky}
Baroni, M., Bernardini, S., et~al. (2006).
\newblock {\em WaCky! working papers on the web as corpus}.
\newblock Gedit.

\bibitem[Baroni et~al., 2009]{baroni2009wacky}
Baroni, M., Bernardini, S., Ferraresi, A., and Zanchetta, E. (2009).
\newblock The wacky wide web: a collection of very large linguistically
  processed web-crawled corpora.
\newblock {\em Language resources and evaluation}, 43:209--226.

\bibitem[Brown et~al., 2020]{brown2020}
Brown, T., Mann, B., Ryder, N., Subbiah, M., Kaplan, J.~D., Dhariwal, P.,
  Neelakantan, A., Shyam, P., Sastry, G., Askell, A., Agarwal, S.,
  Herbert-Voss, A., Krueger, G., Henighan, T., Child, R., Ramesh, A., Ziegler,
  D., Wu, J., Winter, C., Hesse, C., Chen, M., Sigler, E., Litwin, M., Gray,
  S., Chess, B., Clark, J., Berner, C., McCandlish, S., Radford, A., Sutskever,
  I., and Amodei, D. (2020).
\newblock Language models are few-shot learners.
\newblock In Larochelle, H., Ranzato, M., Hadsell, R., Balcan, M., and Lin, H.,
  editors, {\em Advances in Neural Information Processing Systems}, volume~33,
  pages 1877--1901. Curran Associates, Inc.

\bibitem[Conneau et~al., 2020]{conneau-etal-2020-unsupervised}
Conneau, A., Khandelwal, K., Goyal, N., Chaudhary, V., Wenzek, G., Guzm{\'a}n,
  F., Grave, E., Ott, M., Zettlemoyer, L., and Stoyanov, V. (2020).
\newblock Unsupervised cross-lingual representation learning at scale.
\newblock In {\em Proceedings of the 58th Annual Meeting of the Association for
  Computational Linguistics}, pages 8440--8451, Online. Association for
  Computational Linguistics.

\bibitem[Crespo et~al., 2023]{ramos2023carolina}
Crespo, M. C. R.~M., de~Souza Jeannine~Rocha, M.~L.,
  Louren{\c{c}}o~Sturzeneker, M., Ribas~Serras, F., Lamartine~de Mello, G.,
  Silva~Costa, A., Palma, M.~F., Morais~Mesquita, R., Guets, R. d.~P.,
  Marques~da Silva, M., et~al. (2023).
\newblock Carolina: a general corpus of contemporary brazilian portuguese with
  provenance, typology and versioning information.
\newblock {\em arXiv e-prints}, pages arXiv--2303.

\bibitem[Devlin et~al., 2018]{devlin2018bert}
Devlin, J., Chang, M.-W., Lee, K., and Toutanova, K. (2018).
\newblock Bert: Pre-training of deep bidirectional transformers for language
  understanding.
\newblock {\em arXiv preprint arXiv:1810.04805}.

\bibitem[Fonseca et~al., 2016]{fonseca2016assin}
Fonseca, E., Santos, L., Criscuolo, M., and Aluisio, S. (2016).
\newblock Assin: Avaliacao de similaridade semantica e inferencia textual.
\newblock In {\em Computational Processing of the Portuguese Language-12th
  International Conference, Tomar, Portugal}, pages 13--15.

\bibitem[He et~al., 2020]{He2020deberta}
He, P., Liu, X., Gao, J., and Chen, W. (2020).
\newblock Deberta: Decoding-enhanced bert with disentangled attention.
\newblock {\em arXiv preprint arXiv:2006.03654}.

\bibitem[Izsak et~al., 2021]{Izsak2021bert}
Izsak, P., Berchansky, M., and Levy, O. (2021).
\newblock How to train {BERT} with an academic budget.
\newblock In {\em Proceedings of the 2021 Conference on Empirical Methods in
  Natural Language Processing}, pages 10644--10652. Association for
  Computational Linguistics.

\bibitem[Jaderberg et~al., 2017]{Jaderberg2017pbt}
Jaderberg, M., Dalibard, V., Osindero, S., Czarnecki, W.~M., Donahue, J.,
  Razavi, A., Vinyals, O., Green, T., Dunning, I., Simonyan, K., Fernando, C.,
  and Kavukcuoglu, K. (2017).
\newblock Population based training of neural networks.
\newblock {\em arXiv preprint arXiv:1711.09846}.

\bibitem[Joshi et~al., 2020]{Joshi2020stateFate}
Joshi, P., Santy, S., Budhiraja, A., Bali, K., and Choudhury, M. (2020).
\newblock The state and fate of linguistic diversity and inclusion in the {NLP}
  world.
\newblock In {\em Proceedings of the 58th Annual Meeting of the Association for
  Computational Linguistics}, pages 6282--6293, Online. Association for
  Computational Linguistics.

\bibitem[Kaplan et~al., 2020]{Kaplan2020scaling}
Kaplan, J., McCandlish, S., Henighan, T., Brown, T.~B., Chess, B., Child, R.,
  Gray, S., Radford, A., Wu, J., and Amodei, D. (2020).
\newblock Scaling laws for neural language models.
\newblock {\em arXiv preprint arXiv:2001.08361}.

\bibitem[Liu et~al., 2019]{liu2019roberta}
Liu, Y., Ott, M., Goyal, N., Du, J., Joshi, M., Chen, D., Levy, O., Lewis, M.,
  Zettlemoyer, L., and Stoyanov, V. (2019).
\newblock Roberta: A robustly optimized bert pretraining approach.
\newblock {\em arXiv preprint arXiv:1907.11692}.

\bibitem[R.~Serras and Sturzeneker, 2023]{slides_carolina12}
R.~Serras, F. and Sturzeneker, M. (2023).
\newblock Carolina: um córpus geral do português brasileiro com
  proveniência, tipologia e versionamento.
\newblock note = {Accessed: {2023–09-18}}.

\bibitem[Raffel et~al., 2020]{raffel2020}
Raffel, C., Shazeer, N., Roberts, A., Lee, K., Narang, S., Matena, M., Zhou,
  Y., Li, W., and Liu, P.~J. (2020).
\newblock Exploring the limits of transfer learning with a unified text-to-text
  transformer.
\newblock {\em J. Mach. Learn. Res.}, 21(1).

\bibitem[Real et~al., 2020]{real2020assin}
Real, L., Fonseca, E., and Gon{\c{c}}alo~Oliveira, H. (2020).
\newblock The assin 2 shared task: a quick overview.
\newblock In {\em Computational Processing of the Portuguese Language: 14th
  International Conference, PROPOR 2020, Evora, Portugal, March 2--4, 2020,
  Proceedings 14}, pages 406--412. Springer.

\bibitem[Rodrigues et~al., 2023]{rodrigues2023advancing}
Rodrigues, J., Gomes, L., Silva, J., Branco, A., Santos, R., Cardoso, H.~L.,
  and Os{\'o}rio, T. (2023).
\newblock Advancing neural encoding of portuguese with transformer albertina
  pt.
\newblock {\em arXiv preprint arXiv:2305.06721}.

\bibitem[Souza et~al., 2020]{souza2020bertimbau}
Souza, F., Nogueira, R., and Lotufo, R. (2020).
\newblock Bertimbau: pretrained bert models for brazilian portuguese.
\newblock In {\em Intelligent Systems: 9th Brazilian Conference, BRACIS 2020,
  Rio Grande, Brazil, October 20--23, 2020, Proceedings, Part I 9}, pages
  403--417. Springer.

\bibitem[Sturzeneker et~al., 2022]{sturzeneker2022carolina}
Sturzeneker, M., Crespo, M.~C., Rocha, M.~L., Finger, M., de~Sousa, M. C.~P.,
  do~Monte, V.~M., and Namiuti, C. (2022).
\newblock Carolina's methodology: building a large corpus with provenance and
  typology information.
\newblock In {\em DHandNLP@ PROPOR}, pages 53--58.

\bibitem[Tay et~al., 2022]{Tay2022scale}
Tay, Y., Dehghani, M., Rao, J., Fedus, W., Abnar, S., Chung, H.~W., Narang, S.,
  Yogatama, D., Vaswani, A., and Metzler, D. (2022).
\newblock Scale efficiently: Insights from pretraining and finetuning
  transformers.
\newblock In {\em International Conference on Learning Representations}.

\bibitem[Vargas et~al., 2022]{vargas-etal-2022-hatebr}
Vargas, F., Carvalho, I., Rodrigues~de G{\'o}es, F., Pardo, T., and Benevenuto,
  F. (2022).
\newblock {H}ate{BR}: A large expert annotated corpus of {B}razilian
  {I}nstagram comments for offensive language and hate speech detection.
\newblock In {\em Proceedings of the Thirteenth Language Resources and
  Evaluation Conference}, pages 7174--7183, Marseille, France. European
  Language Resources Association.

\bibitem[Vaswani et~al., 2017]{vaswani2017attention}
Vaswani, A., Shazeer, N., Parmar, N., Uszkoreit, J., Jones, L., Gomez, A.~N.,
  Kaiser, {\L}., and Polosukhin, I. (2017).
\newblock Attention is all you need.
\newblock {\em Advances in neural information processing systems}, 30.

\bibitem[Wagner~Filho et~al., 2018]{wagner2018brwac}
Wagner~Filho, J.~A., Wilkens, R., Idiart, M., and Villavicencio, A. (2018).
\newblock The br{W}a{C} corpus: A new open resource for {B}razilian
  {P}ortuguese.
\newblock In {\em Proceedings of the Eleventh International Conference on
  Language Resources and Evaluation ({LREC} 2018)}, Miyazaki, Japan. European
  Language Resources Association (ELRA).

\bibitem[Xue et~al., 2021]{xue-etal-2021-mt5}
Xue, L., Constant, N., Roberts, A., Kale, M., Al-Rfou, R., Siddhant, A., Barua,
  A., and Raffel, C. (2021).
\newblock m{T}5: A massively multilingual pre-trained text-to-text transformer.
\newblock In {\em Proceedings of the 2021 Conference of the North American
  Chapter of the Association for Computational Linguistics: Human Language
  Technologies}, pages 483--498, Online. Association for Computational
  Linguistics.

\end{thebibliography}

\end{document}